\documentclass[final,3p,times,twocolumn, authoryear]{elsarticle}
\usepackage{framed,multirow}

\usepackage{url}
\usepackage{xcolor}
\definecolor{newcolor}{rgb}{.8,.349,.1}

\usepackage{pdfpages}

\usepackage{amssymb}

\usepackage{amsmath,amsfonts,amssymb}
\usepackage{textcomp}
\usepackage{gensymb}
\usepackage{hyperref}

\usepackage{booktabs}
\RequirePackage{geometry}
\usepackage{xcolor}

\usepackage{wrapfig}
\usepackage{float}

\usepackage{graphicx}

\usepackage[font=small,belowskip=6pt]{subcaption} 
\usepackage[font=footnotesize,labelfont=bf]{caption}
\usepackage{float}
\usepackage[normalem]{ulem}

\usepackage{array}
\newcolumntype{H}{>{\setbox0=\hbox\bgroup}c<{\egroup}@{}}

\newif\ifdoubleblind

\newcommand{\authorbio}[3]
{
{
    \begin{wrapfigure}{l}{25mm} 
        \vspace{-5mm}
        \includegraphics[width=1.1in]{#3}
    \end{wrapfigure}\par
    \noindent \textbf{#1} #2 \par
}
}
\newcommand{\cocometricstable}[4][AP]
{
\begin{table*}[ht]
    \centering
    \caption{#3}
    \vspace*{-3mm}
    \begin{tabular}{l|ccc|ccc}
     & \bfseries $\text{#1}$ & \bfseries $\text{#1}_{50}$ & \bfseries $\text{#1}_{75}$ & \bfseries $\text{#1}_{S}$ & \bfseries $\text{#1}_{M}$ & \bfseries $\text{#1}_{L}$ \\
    \hline
    \hline
    #4
    \end{tabular}
    \label{#2}
\end{table*}
}
\newcommand{\cocometricstableshape}[4][AP]
{
\begin{table*}[ht]
    \centering
    \caption{#3}
    \vspace*{-3mm}
    \begin{tabular}{ll|ccc|cccH}
     Mask shape & &\bfseries $\text{#1}$ & \bfseries $\text{#1}_{50}$ & \bfseries $\text{#1}_{75}$ & \bfseries $\text{#1}_{S}$ & \bfseries $\text{#1}_{M}$ & \bfseries $\text{#1}_{L}$ &  \bfseries SU \\
    \hline
    \hline
    #4
    \end{tabular}
    \label{#2}
\end{table*}
}
\newcommand{\cocometricstablespeedup}[4][AP]
{
\begin{table*}[ht]
    \centering
    \caption{#3}
    \vspace*{-3mm}
    \begin{tabular}{l|ccc|cccH}
     & \bfseries $\text{#1}$ & \bfseries $\text{#1}_{50}$ & \bfseries $\text{#1}_{75}$ & \bfseries $\text{#1}_{S}$ & \bfseries $\text{#1}_{M}$ & \bfseries $\text{#1}_{L}$ &  \bfseries SU\\
    \hline
    \hline
    #4
    \end{tabular}
    \label{#2}
\end{table*}
}

\journal{Computer Vision and Image Understanding, \url{http://doi.org/10.1016/j.cviu.2019.102795}}

\usepackage{etoolbox}
\makeatletter
\patchcmd{\ps@pprintTitle}
  {Preprint submitted to}
  {Published in}
  {}{}
\makeatother

\begin{document}

\begin{frontmatter}

\title{Faster Training of Mask R-CNN by Focusing on Instance Boundaries\tnoteref{fn:equalcontribution}}

\address[autadd:bmw]{BMW Car IT GmbH, Lise-Meitner-Stra\ss e 14, 89081 Ulm, Germany}
\address[autadd:goe]{Georg-August University of G\"ottingen, Friedrich-Hund-Platz 1, 37077  G\"ottingen, Germany}
\address[autadd:frei]{Albert Ludwig University of Freiburg, Fahnenbergplatz, 79085 Freiburg im Breisgau, Germany}

\tnotetext[fn:equalcontribution]{Both authors contributed equally to this work and must both be cited. They are both corresponding authors.}

\author[autadd:bmw,autadd:goe,fn:rsz]{Roland S. Zimmermann}
\fntext[fn:rsz]{R. S. Zimmermann is with the University of G\"ottingen, G\"ottingen, Germany. This work was started when he was an intern at BMW Car IT GmbH, Ulm, Germany. Email: roland.zimmermann@stud.uni-goettingen.de}
\author[autadd:bmw,autadd:frei,fn:jns]{Julien N. Siems}
\fntext[fn:jns]{J. N. Siems is with the Albert Ludwig University of Freiburg, Freiburg, Germany. This work was started when he was an intern at BMW Car IT GmbH, Ulm, Germany. Email: siemsj@cs.uni-freiburg.de}

\begin{abstract}
    We present an auxiliary task to Mask R-CNN, an instance segmentation network, which leads to faster training of the mask head. Our addition to Mask R-CNN is a new prediction head, the Edge Agreement Head, which is inspired by the way human annotators perform instance segmentation. Human annotators copy the contour of an object instance and only indirectly the occupied instance area. Hence, the edges of instance masks are particularly useful as they characterize the instance well. The Edge Agreement Head therefore encourages predicted masks to have similar image gradients to the ground-truth mask using edge detection filters. We provide a detailed survey of loss combinations and show improvements on the MS COCO Mask metrics compared to using no additional loss. Our approach marginally increases the model size and adds no additional trainable model variables. While the computational costs are increased slightly, the increment is negligible considering the high computational cost of the Mask R-CNN architecture. As the additional network head is only relevant during training, inference speed remains unchanged compared to Mask R-CNN. In a default Mask R-CNN setup, we achieve a training speed-up and a relative overall improvement of 8.1\% on the MS COCO metrics compared to the baseline.
\end{abstract}

\end{frontmatter}


\section{Introduction}
    Significant improvements in computer vision techniques have been made possible by the rapid progress of training Deep Convolutional Neural Networks in recent years. Application areas include image classification \citep{krizhevsky2012imagenet, szegedy2015going, Simonyan15, he2016deep} and object detection \citep{girshick2015fast, redmon2016you, liu2016ssd}. One of the most demanding computer vision tasks is instance segmentation, as it involves localizing and segmenting object instances. Recently, there have been multiple methods \citep{Li2017FullyCI, bai2017deep, liu2018path, he2017mask} proposed to perform this task.
    
    Another beneficial factor to the success of these Deep Learning architectures is the availability of large labeled datasets such as MS COCO \citep{lin2014microsoft} and the Cityscapes dataset \citep{cordts2016cityscapes}. Labeling an image dataset for instance segmentation is particularly time-consuming, because it requires segmenting all objects in a scene. It is therefore highly desirable to speed up training of an instance segmentation model to be more data efficient. In this work, we propose a conceptually straightforward addition to the Mask R-CNN \citep{he2017mask} architecture which reduces training time of the mask branch.\\
    
    The Mask R-CNN architecture is based on Faster R-CNN \citep{ren2015faster}, which introduced an efficient Region Proposal Network (RPN) design to output bounding box proposals.
    The proposals are computed using a sliding window approach to make them translation invariant. A feature extractor such as ResNet \citep{he2016deep}, Inception \citep{szegedy2017inception} or VGGNet \citep{Simonyan15} is used as input to the region proposal network. The regions and features are used in the bounding box regression head, that refines the bounding box localization and the softmax classification head, which determines the instance class. This second stage is the architecture as described in Fast R-CNN \citep{girshick2015fast}.
    
    Mask R-CNN is a simple but effective addition to the Faster R-CNN architecture that adds a head for instance mask prediction. Using a small Fully Convolutional Neural Network (FCN) \citep{long2015fully}, it can predict pixel level instance masks. Besides the mask branch, it uses a Feature Pyramid Network (FPN) backbone as proposed by \cite{lin2017feature}. This addition allows the network to make use of both high-resolution feature maps in the lower layers for accurate localization, as well as semantically more meaningful higher-level features, which are of lower resolution. 
    Another contribution is ROI Align which maps arbitrarily sized spatial regions of interest in the features to a fixed spatial resolution using bilinear interpolation. This modification improves the COCO Mask metrics and enables the use of instance masks which require precise localization. 
    
    For the mask head, a new loss term $L_{Mask}$ has been introduced, which calculates the pixel-wise cross entropy between the predicted and target masks. The Mask R-CNN loss function
    \begin{align}
        \label{eq:mrcnn_loss}
        L_{MRCNN} = L_{Class} + L_{Box} + L_{Mask} 
    \end{align}
    is a multi-task loss based on the Faster R-CNN loss.
    
    We propose to attach an Edge Agreement Head to the mask branch of Mask R-CNN which acts as an auxiliary task to Mask R-CNN. This head uses traditional edge detection filters such as Sobel and Laplacian kernels \citep{Sobel1973, Ponce2012} on both the predicted mask and the ground-truth mask to encourage their edges to agree. Instances in natural images are bounded by the edges that annotators use to mark the instance. Therefore, we show that encouraging the edges in the predicted and ground-truth mask to agree leads to faster training of our mask head. We argue that this is a result of the instance boundary being a robust feature to mask prediction, which can be easily propagated from the image to the mask branch. 

\section{Related Work}
    \textbf{Multi Task Learning.} The Edge Agreement Head acts as an auxiliary task \citep{Ruder2017AnOO} to the multi-task model Mask R-CNN, which is performing both object detection and instance segmentation. Auxiliary tasks have shown to encourage models to learn robust representations of their input in a variety of applications, such as facial landmark detection \citep{zhang2014facial}, natural language processing \citep{collobert2008unified} or steering prediction in autonomous driving \citep{Caruana:1997:ML:262868.262872}. Even seemingly unrelated tasks, e.g. weather prediction to semantic scene segmentation, can improve the model's overall performance \citep{liebel2018auxiliary}. \\
    
    \textbf{Monocular Depth Estimation.} \cite{godard2017unsupervised} use image gradients to encourage consistency between input images and predicted disparity maps. However, the left-right disparity consistency loss does not ensure image gradients of the predicted disparity of the left and right camera to exhibit similar edge detection filter responses. \\
    
    \textbf{Scene segmentation.} \cite{chen2016semantic} show a two-part model predicting both semantic segmentations and edges. The semantic segmentation model is based on the DeepLab model \citep{Chen2018DeepLabSI} and the edge detection filter is created using intermediate convolutional filters of the DeepLab model. The task specific edge-detection on the input image is used to refine the coarse segmentation using domain transform. Our approach determines edges in fixed size, low dimensional  instance mask images, for which traditional edge detection filters have been proven to be effective. 
    Similarly, \cite{marmanis2018classification} predict both semantic scene segmentation and semantic boundaries. The network responsible for predicting semantic boundaries is trained using a Euclidean loss before each pooling layer  to enforce each layer to predict edges at different scales. Our approach uses predefined edge detection filters with well-known properties, which are kept constant during training, leaving us with a significantly lower additional memory footprint and computational costs. \\

    \textbf{Edge detection.} The detection of edges has been a research topic for many decades and numerous methods have been proposed \citep{Sobel1973, konishi2003statistical}. This field has seen large improvements due to deep learning techniques \citep{bertasius2015deepedge, shen2015deepcontour, xie2015holistically}. Our work uses the Sobel image gradient filters proposed in \citep{Sobel1973}, because it keeps the computational overhead to a minimum. Furthermore, our edge detection filters are used on $28 \times 28$ sized masks with only one channel depicting a single instance and not high-resolution color images. This significantly reduces the complexity of the problem and justifies our choice of simple edge detection filters.\\

    \textbf{Instance segmentation.} \cite{hayder2017boundary} propose a model that predicts the truncated distance transform \citep{borgefors1986distance} of the mask, making it more resilient towards non-instance enclosing bounding box proposals. The proposed architecture for the boundary-aware instance segmentation network has many similarities to Mask R-CNN as they are both based on the Faster R-CNN architecture by \cite{ren2015faster}. However, they achieved lower results on the instance segmentation benchmark on the Cityscapes dataset compared to Mask R-CNN, which we are basing our work on. Yang et al. \citep{yang2016object} propose an encoder - decoder architecture which predicts object contours. For training on the MS COCO dataset \citep{lin2014microsoft} the coarse polygon ground-truth edges are refined to follow the object contours more closely by applying a dense conditional random field \citep{krahenbuhl2011efficient} or applying graph cut \citep{boykov2001interactive}. This is particularly necessary in this case since the model predicts high resolution edges. Since the instance masks by Mask R-CNN only have a low resolution of e.g. 28 x 28, we argue that the preprocessing is unnecessary in this case, since most details are lost at this scale. It was also not used by the original Mask R-CNN Paper by \cite{he2017mask}. Kirillov et al. \citep{kirillov2017instancecut} propose a model combination which predicts both a semantic segmentation and edges. The output of the edge score is used to compute super pixels which, alongside the predictions and the edge score, are used to solve a Multi Cut problem \citep{chopra1993partition} which predicts instances. This work is different from our experiments since we first find instance masks and then compare the predicted and ground-truth edges of these. In addition, this work operates on high resolution images, which complicates training, since the ground-truth edges only occupy a very small region of the overall area. The authors therefore rescale the cost function for underrepresented classes. This step is less relevant for the Edge Agreement Head applied to Mask R-CNN since the masks are predicted based on a proposed bounding box and because the class and mask prediction are decoupled in the Mask R-CNN architecture \citep{he2017mask}.

\section{Edge Agreement Loss}
\begin{figure}
        \centering
        \vspace*{-1mm}
        \includegraphics[width=0.35\textwidth]{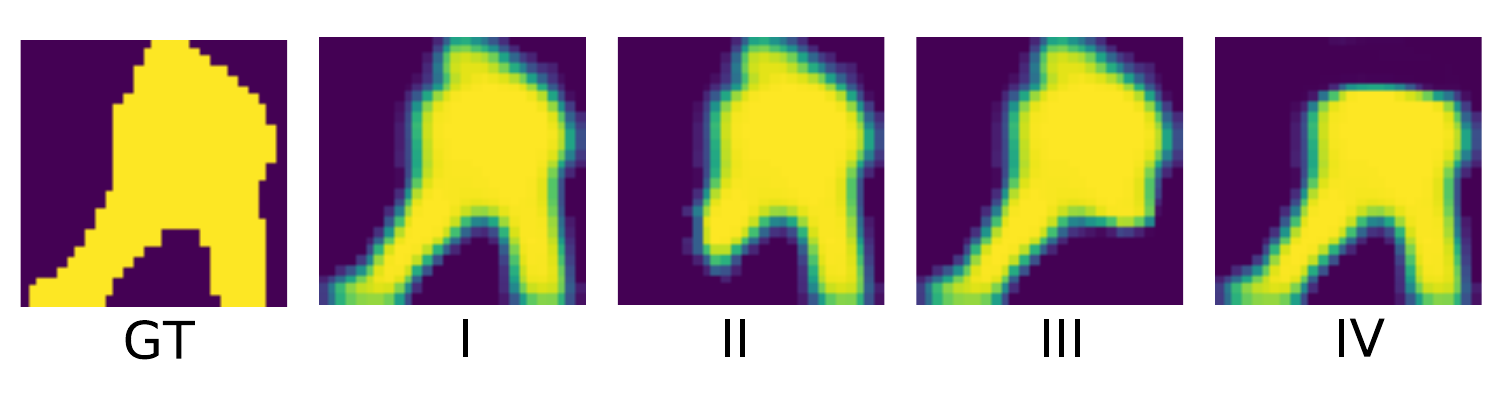}
        \vspace*{-3mm}
        \caption{Overview of different example masks to illustrate the effect of the Edge Agreement Loss. \textit{GT} corresponds to the ground truth  and \textit{I} to \textit{IV} represent four example mask predictions which demonstrate early-stage predictions of the Mask R-CNN during training.}
        \label{fig:mask_gt_prediction_overview}
    \end{figure}

    When training a Mask R-CNN for instance segmentation one often observes incomplete or poor masks, especially during early training steps. Furthermore, the masks often do not follow the real object boundaries. Possible mistakes such as missing parts or oversegmentation are illustrated in Figure~\ref{fig:mask_gt_prediction_overview}.
    
    To reduce this problem, we draw our inspiration from how a human would perform instance segmentation: instead of immediately assigning parts of the image to specific objects one often identifies at first the boundaries of the object and fills the enclosed area. To help the network perform the segmentation in an analogous way, i.e. show the importance of edges and boundaries of objects, we have constructed an auxiliary loss called Edge Agreement Loss $L_{Edge}$. It is defined as the $L^p$ loss between the edges in the predicted mask and the ground-truth mask. The total loss $L_{Total}$ consists of the original Mask R-CNN loss $L_{MRNN}$ (eq. \ref{eq:mrcnn_loss}) and the new Edge Agreement Loss $L_{Edge}$ which are summed. To compute this new loss, the first step is to identify the edges in the predicted and the ground-truth mask.
    
\subsection{Edge Detection}
    In detail, we examined edge detection filters which can be described as a convolution with a $3\times3$ kernel, such as the well-known \textit{Sobel} and \textit{Laplacian} filters.\\

    The Sobel filters \citep{Sobel1973} are two-dimensional filters to detect edges. As the filters describe a first-order gradient operation they are rotation-dependent. There are two filters
    \begin{align}
        S_x = \begin{bmatrix}
        1 & 0 & -1 \\
        2 & 0 & -2 \\
        1 & 0 & -1 \\
    \end{bmatrix}, \quad
        S_y =\begin{bmatrix}
        1 & 2 & 1 \\
        0 & 0 & 0 \\
        -1 & -2 & -1 \\
        \end{bmatrix}
    \end{align}
    which describe the horizontal and the vertical gradient respectively. An edge in the image corresponds to a high absolute response along the filter's direction. In the following the concatenation of both filters into a $3 \times 3 \times 2$ dimensional tensor is referenced as the Sobel filter $S$.\\

    The Laplacian filter is a discretization of the two-dimensional Laplacian operator (i.e. the second derivative). The filter
    \begin{align}
        L =\begin{bmatrix}
        0 & 1 & 0 \\
        1 & -4 & 1 \\
        0 & 1 & 0 \\
        \end{bmatrix}
    \end{align}
    is the direct result of a finite-difference approximation of the derivative \citep{Ponce2012}. The operator is known to be rotation invariant which means that it can detect edges in both x and y direction. As it is a second-order operator, an edge in the image corresponds to a zero-crossing, rather than a strong filter response. By including the main- and anti-diagonal elements the filter can be made responsive to 45\degree \ angles
    \begin{align}
        L =\begin{bmatrix}
        1 & 1 & 1 \\
        1 & -8 & 1 \\
        1 & 1 & 1 \\
        \end{bmatrix}.
    \end{align}
    
    This is the Laplacian kernel ($L$) used in all further experiments. 
    
    In addition to these exemplary kernels we also used the Prewitt operator \citep{lipkin1970picture}, Kayyali filter \citep{kawalec2014edge} and the Roberts operator \citep{roberts1963machine}.
    
    \begin{figure}[h]
        \centering
        \includegraphics[height=2.65cm]{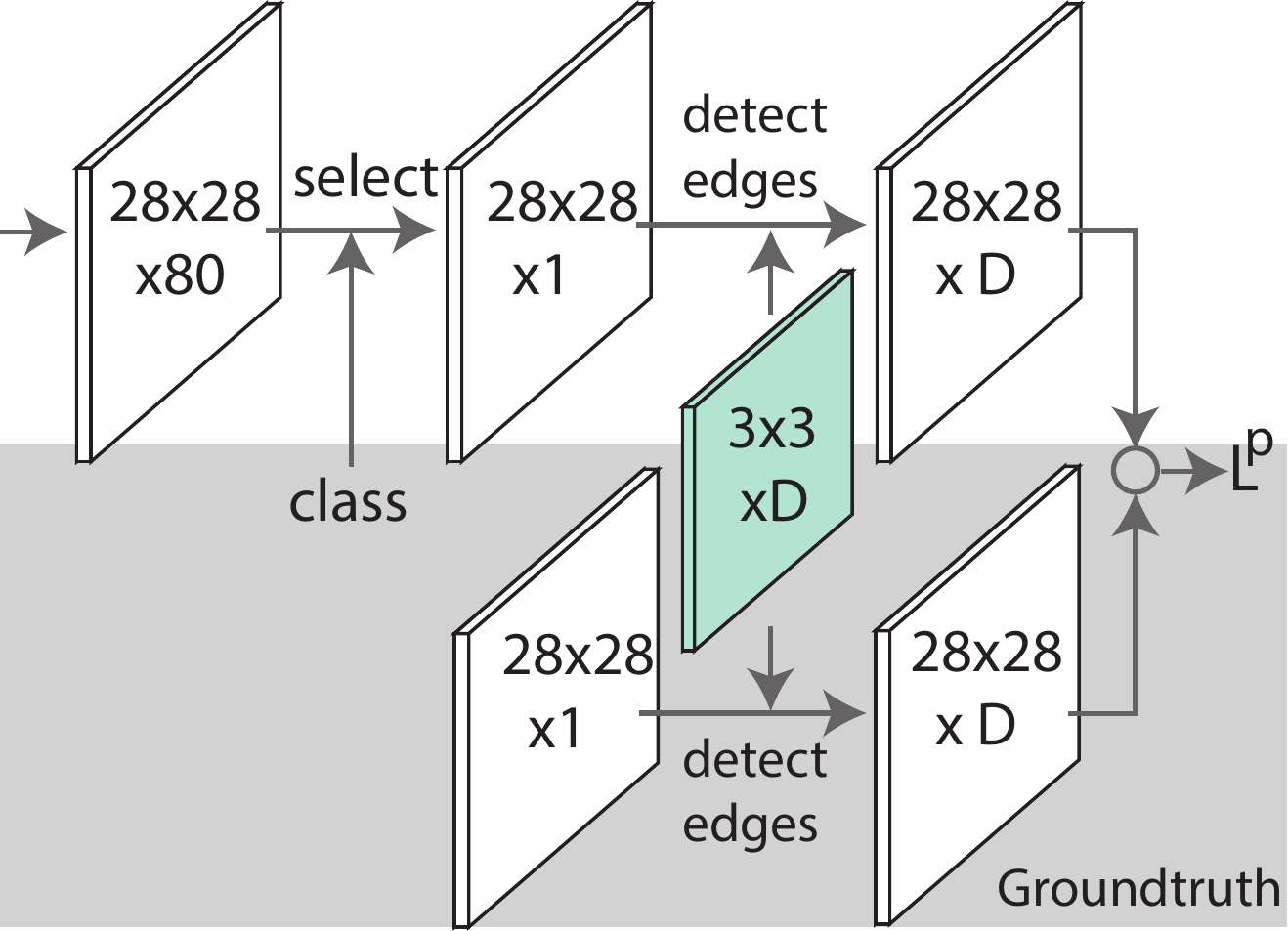}
        \vspace*{-1mm}
        \caption{Edge Agreement Head: We extend the existing mask branch architecture. Of the $28 \times 28 \times 80$ dimensional output of the mask branch, the mask corresponding to the correct class is selected. The head computes a convolution of the selected mask and the ground-truth mask with the $3 \times 3 \times D$ dimensional edge detection filter (turquoise). Between these a $L^p$ loss is calculated, which results in the term $L_{Edge}$ (Best viewed in color).}
        \label{fig:edge_loss_head}
    \end{figure}
    
    \begin{figure}[h]
        \centering
        \vspace*{-2mm}
        \includegraphics[width=3.4in]{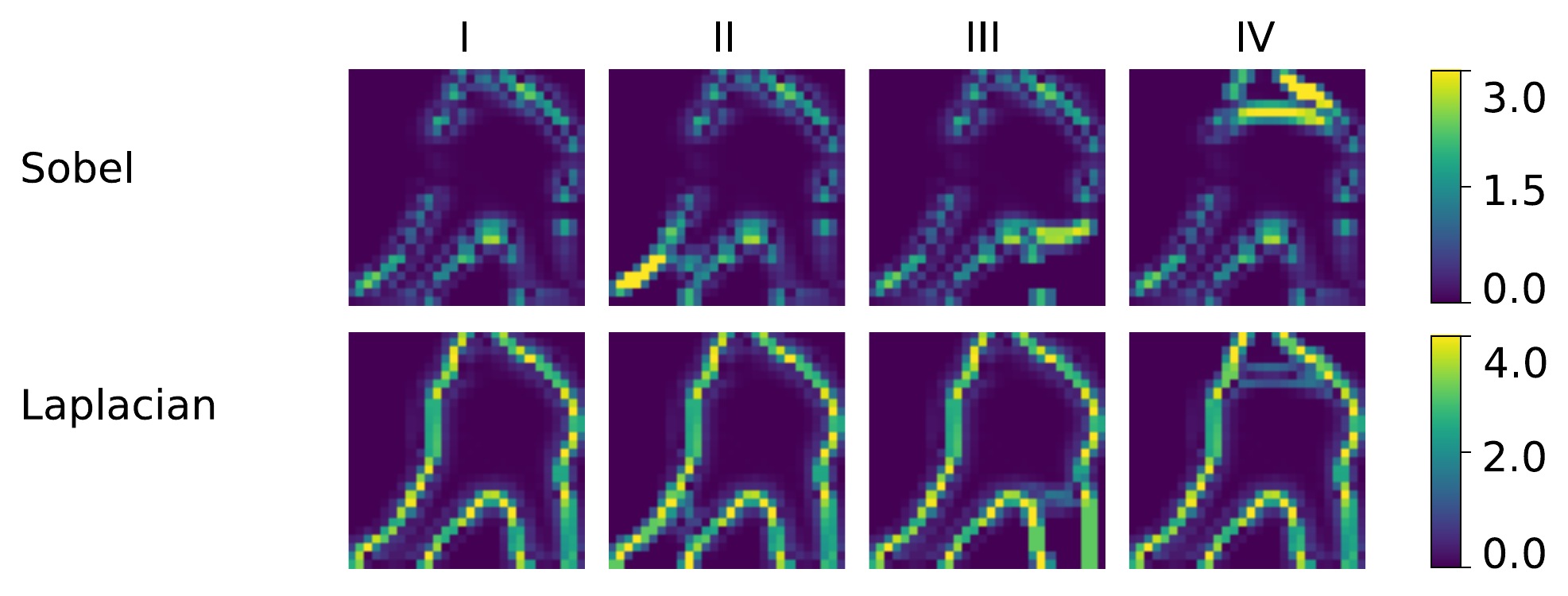}
        \vspace*{-2mm}
        \caption{$L^2$ errors for the four example predictions and the different methods. Each column $I$ to $IV$ corresponds to one of the examples in Figure~\ref{fig:mask_gt_prediction_overview}. The first row shows the $L^2$ error based on the Sobel filter magnitude, while for the second row the Laplace filter is used.}
        \label{fig:mask_gt_prediction_blurred_gt_loss}
    \end{figure}
    
\subsection{Loss Construction}
    \label{sec:loss_construction}
    To calculate the final loss $L_{Edge}$ we propose an additional network head, called the Edge Agreement Head. It uses the predicted and the matched ground-truth masks as input, which are then convolved with a selection of edge detection filters. Afterwards, the difference between the predicted and ground-truth edge maps are determined. The entire procedure is illustrated in Figure~\ref{fig:edge_loss_head} in the left half. For this task, we choose the set of $L^p$ loss functions. Mathematically they can be expressed as the $p$-th power of the generalized power mean $M_p$ of the absolute difference between the target $\hat{\mathbf{y}}$ and the prediction $\mathbf{y}$
    \begin{align}
        L^p(\mathbf{y}, \hat{\mathbf{y}}) = M_p(|\mathbf{y} - \hat{\mathbf{y}}|)^p.
    \end{align}
    For $p=2$ this equals the mean square error, commonly used in deep learning.\\
    
    The edge agreement head can be calculated with only minimal additional computational and memory requirements. This means, that the method can be integrated in existing systems for training Mask R-CNN without requiring new or additional hardware.
    
\section{Implementation Details}\label{sec:implementation}
    A mask size of $28\times28$ pixels and an image resolution of $1024\times800$ pixels are used. All training images are resized to this size preserving their aspect ratio. As the training images may have different aspect ratios, the remaining space of the image is zero padded. This method differs from the one used in the original Mask R-CNN implementation \citep{he2017mask}, where resizing is done such that the smallest side is $800$ pixels and the largest is trimmed at $1000$ pixels.\\ 
    
    The ResNet \citep{he2016deep} feature extractor is initialized with weights trained on ImageNet \citep{deng2009imagenet}; all other weights (e.g. in the region proposal network) are initialized using Xavier initialization \citep{glorot2010understanding}. \\
    
    A similar training strategy to other Mask R-CNN work \citep{he2017mask} is followed. We choose to train the network for 160k steps on the MS COCO 2017 train dataset with a batch size of 2 on a single GPU machine, while for Mask R-CNN an effective batch size of 16 was used. The training consists of three stages each lasting for 40k, 80k, 40k steps respectively: in the first stage only the Mask R-CNN branches and not the ResNet backbone are trained. Next, the prediction heads and parts of the backbone (starting at layer 4) are optimized. Finally, in the third stage, the entire model (backbone and heads) is trained together. For the first two training stages we use a learning rate of $0.001$ and for the last one a decreased learning rate of $0.001/10$. The optimization is done by SGD with momentum set to $0.9$ and weight decay set to $0.0001$.

\section{Experiments}
    We perform our experiments using the implementation of Mask R-CNN by matterport \citep{matterport_maskrcnn_2017}, based on the \textsc{Keras} framework \citep{chollet2015keras} with a \textsc{TensorFlow} backend \citep{tensorflow2015-whitepaper}. Each training is carried out on a single GPU using either an NVIDIA Titan X or an NVIDIA GeForce GTX 1080 Ti.\\
    
    We examine three aspects of the proposed Edge Agreement Head. At first, we inspect the influence of the edge detection filters on the training speed (section \ref{sec:influence_of_filters}). In section \ref{sec:lp_loss}, the different metrics of the $L^p$ family are used to examine the influence of the loss function's steepness on the training speed. Section \ref{sec:lp_loss_wf_factor} shows the impact weighting the Edge Agreement Loss has on the overall loss. We investigate the influence of the Edge Agreement Head with varied mask size in section \ref{sec:influence_of_mask_size}. In section \ref{sec:longer_training} we show the results on the metrics after longer training. Finally, in section \ref{sec:other_experiments} we elaborate on modifications to the Edge Agreement Head which did not have a positive effect on the training.
    
    For all experiments, we follow the same scheme: every network configuration examined is trained and evaluated three times. The resulting training curves and metrics displayed are the averaged values. Furthermore, to be able to compare all runs and to reduce the time required for all experiments we do not train the networks until they have converged, but only for a fixed and limited number of training steps. The only data augmentation used in all three steps are random horizontal flips.
    
    We present the COCO metrics for our experiments and compare them with the results obtained using a Mask R-CNN model without modifications (baseline) for every experiment conducted. A significant disadvantage of using the COCO Mask metrics is that they do not compare the ground-truth and predicted mask pixel per pixel since they only consider the area and the instance enclosing bounding box. As a result, the COCO Mask metrics are unsuitable to compare the improvement in the details of instance masks which are located on the inside of the mask and do not affect the extremities. We use the COCO metrics because of their dominance in other publications. The mask loss however, can be regarded as a better metric to compare the quality of the instance mask, because it is computed using a cross entropy between matched ground-truth and predicted mask.

\subsection{Influence of Filters}\label{sec:influence_of_filters}
    \begin{figure*}
         \centering
         \vspace*{-3mm}
        \begin{subfigure}{.33\linewidth}
            \centering
            \includegraphics{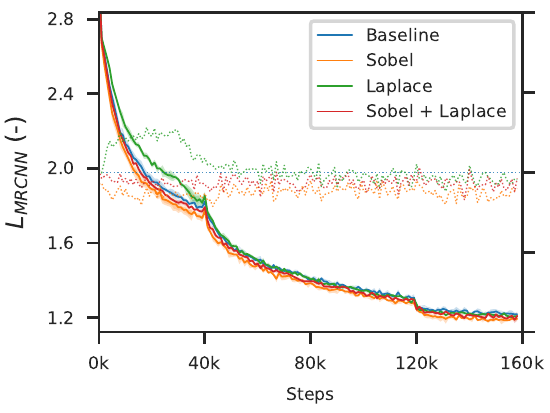}
            \caption{Influence of edge detection filter}\label{fig:influence_filters}
        \end{subfigure}
            \hfill
        \begin{subfigure}{.33\linewidth}
            \centering
            \includegraphics{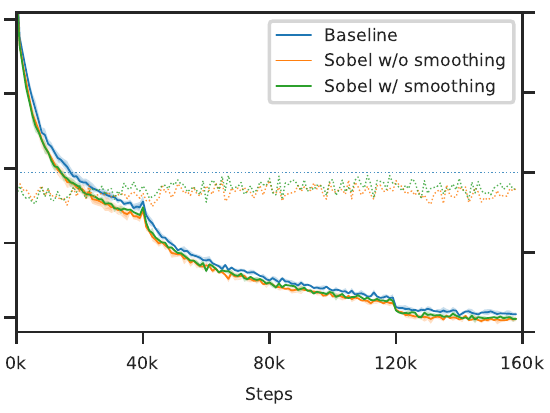}
            \caption{Influence of Gaussian smoothing }\label{fig:influence_smoothing}
        \end{subfigure}
           \hfill
        \begin{subfigure}{.33\linewidth}
            \centering
            \includegraphics{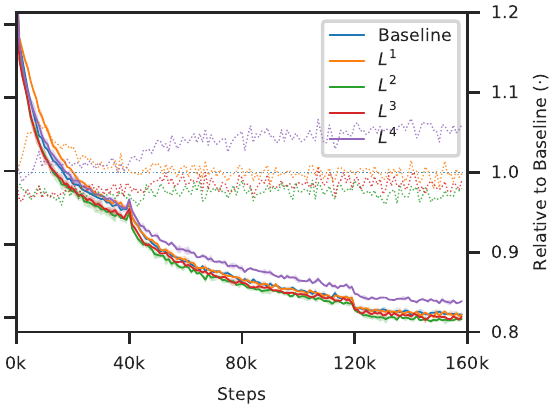}
            \caption{Influence of $p$ in $L^p$ loss}\label{fig:influence_lp}
        \end{subfigure}
        
        \bigskip
        \vspace*{-4mm}

        \begin{subfigure}{.33\linewidth}
            \centering
            \includegraphics{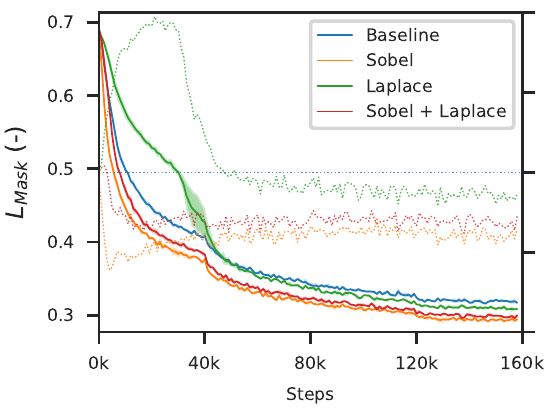}
        \end{subfigure}
            \hfill
        \begin{subfigure}{.33\linewidth}
            \centering
            \includegraphics{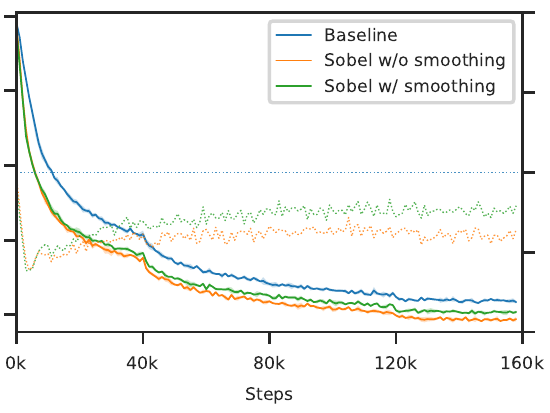}
        \end{subfigure}
           \hfill
        \begin{subfigure}{.33\linewidth}
            \centering
            \includegraphics{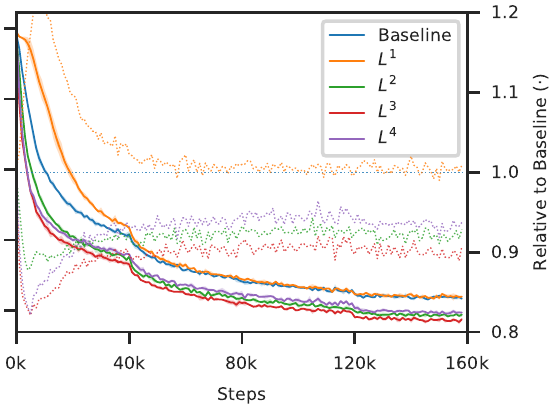}
        \end{subfigure}
        \vspace*{-3mm}
        \label{fig:influence}
        \caption{Comparison of different Edge Agreement Head configurations on the MS COCO dataset. The left y-axis corresponds to the absolute loss values (solid lines) and the right y-axis corresponds to the relative improvement compared to the baseline (dotted lines). The first row shows the original Mask R-CNN Loss $L_{MRCNN}$ while the second row shows the Mask Loss $L_{Mask}$. The first column illustrates the influence of different edge detectors used in the Edge Agreement Head, while the second demonstrates the influence of Gaussian smoothing when using a Sobel edge detection filter (see section \ref{sec:other_experiments}). The last column compares the performance of different $L^p$ loss functions for the Edge Agreement Loss (Best viewed in color).}
    \end{figure*}
    
    \begin{figure*}[!t]
        \vspace{-2mm}
        \centerline{\includegraphics[width=1.0\textwidth]{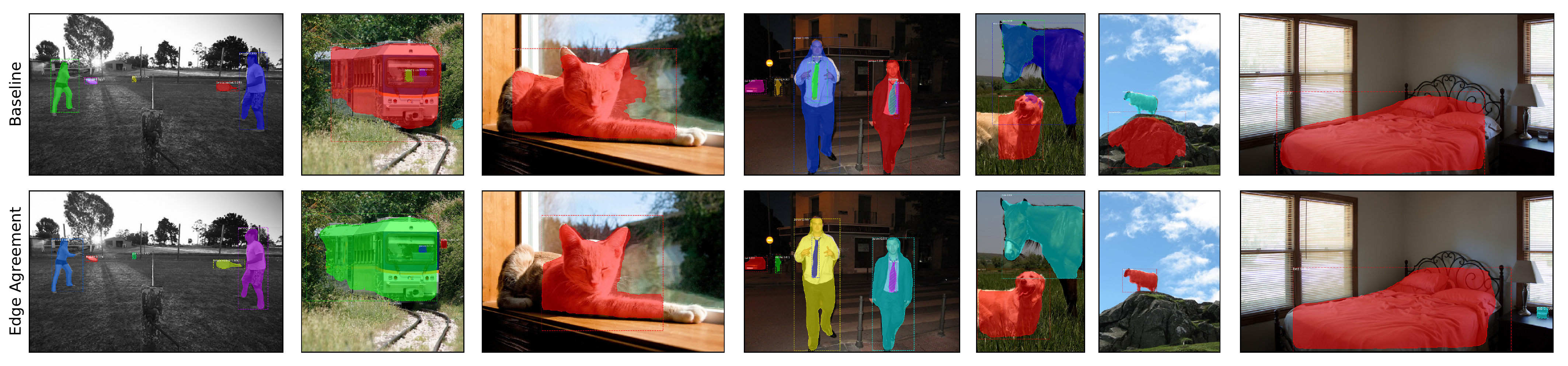}}
        \vspace{-4mm}
        \caption{Comparison between masks predicted by Baseline Mask R-CNN and Mask R-CNN with Edge Agreement Head on the MS COCO dataset using a Sobel edge detection filter after 160k steps on images taken from the MS COCO 2017 dataset \citep{lin2014microsoft} (Best viewed in color).}
        \label{fig:mask_comparison}
    \end{figure*}
    
    In the first experiment the choice of edge detection filters on the training speed and the mask quality is analyzed. The Edge Agreement Loss is computed using the $L^2$ loss. The mask loss $L_{Mask}$ and the original Mask R-CNN loss $L_{MRCNN}$ are displayed in Figure~\ref{fig:influence_filters}. The graphs show that using the Sobel filter leads to a faster decrease of the $L_{MRCNN}$ and the $L_{Mask}$ loss. This is underlined well by plotting $L_{MRCNN}$ while using the Sobel filter relative to the baseline (dotted curves with respect to the right y-axis in Figure~\ref{fig:influence_filters}) which demonstrates a consistent improvement at every training step.
    
    To allow for a comparison between different edge detection filters, we analyzed their performance based on their impact on the other loss terms, as their Edge Agreement Loss magnitude varies depending on the filter. The results on the COCO Mask metrics are shown in Table~\ref{tab:experiment_filters}. Note, that these results are lower than the results reported by \citep{he2017mask}, since we use a different batch size, input image resolution and implementation of Mask R-CNN (\citep{matterport_maskrcnn_2017} not \citep{Detectron2018}). Using the Sobel edge detection filter gave a 7\% relative improvement on the AP score. Notably the Sobel filter resulted in a 12\% relative improvement on the $\text{AP}_{75}$ and a 10\% relative improvement on the $\text{AP}_{S}$ compared to the respective baseline scores. On average, the COCO metrics have been improved relatively by 8.1\%. Using the Laplacian kernel showed only marginal improvements over the baseline. The combination of multiple filters, e.g. Sobel and Laplacian (S \& L), showed no increase in performance. 
    
    A possible explanation for the superiority of the Sobel filter is its structure: as it consists of two filters, not only the strength of an edge along the x and y axis but also the edge's orientation can be used during the gradient descent to minimize the total loss. This additional information accelerates the training. Due to its similar structure, the Prewitt filter showed a comparable effect on the training.
    
    A qualitative comparison between computed masks is shown in Figure~\ref{fig:mask_comparison}. We observe that the models trained with Edge Agreement Loss tend to be less likely to propose bounding boxes which do not contain any object, therefore reducing false positives. This indicates that the features needed to minimize the Edge Agreement Loss are also useful to the Region Proposal Network.
    
    \cocometricstablespeedup[AP]{tab:experiment_filters}{Influence of the choice of edge detection filters on the instance segmentation mask COCO AP metrics on the MS COCO dataset after $160$k steps. Higher is better.}{
        Sobel & $\mathbf{20.2 \pm 0.17}$ & $\mathbf{37.5 \pm 0.37}$ & $\mathbf{20.0 \pm 0.07}$ & $\mathbf{8.8 \pm 0.27}$ & $\mathbf{21.9 \pm 0.18}$ & $\mathbf{28.9 \pm 0.30}$ & $\mathbf{29\%}$ \\
        Prewitt & $20.0 \pm 0.31$ & $\mathbf{37.5 \pm 0.25}$ & $19.6 \pm 0.38$ & $8.5 \pm 0.45$ & $21.6 \pm 0.31$ & $28.1 \pm 0.61$ & $26\%$ \\
        Kayyali & $19.7 \pm 0.16$ & $36.3 \pm 0.25$ & $19.5 \pm 0.18$ & $8.4 \pm 0.32$ & $21.3 \pm 0.14$ & $28.1 \pm 0.46$ & $24\%$ \\
        Roberts & $18.9 \pm 0.31$ & $36.4 \pm 0.44$ & $17.9 \pm 0.28$ & $7.9 \pm 0.21$ & $20.5 \pm 0.41$ & $26.7 \pm 0.53$ & $18\%$\\ 
        Laplace & $19.4 \pm 0.12$ & $36.5 \pm 0.19$ & $18.9 \pm 0.18$ & $8.0 \pm 0.22$ & $21.0 \pm 0.10$ & $27.8 \pm 0.11$ & $21\%$ \\
        S \& L & $20.0 \pm 0.25$ & $37.0 \pm 0.36$ & $19.6 \pm 0.24$ & $8.3 \pm 0.03$ & $21.7 \pm 0.30$ & $28.4 \pm 0.47$ & $23\%$ \\
        \hline
        Baseline & $18.8 \pm 0.14$ & $36.5 \pm 0.24$ & $17.8 \pm 0.13$ & $8.0 \pm 0.21$ & $20.4 \pm 0.29$ & $26.6 \pm 0.24$ & $-$\\
    }
    

    
\subsection{Influence of the Choice $p$ in $L^p$ loss}\label{sec:lp_loss}
    Next, the influence of the exponent $p$ in $L^p$ chosen for the Edge Agreement Loss is analyzed. For all previous experiments, an $L^2$ loss was used; now different values of $p \in \{1,2,3,4\}$ are applied. As an increasing value of $p$ increases the steepness of the loss, falsely detected edges are penalized more strongly. For this evaluation we used the Sobel edge detection filter without smoothing the ground truth.
    
    \cocometricstablespeedup{tab:experiment_lp_norm}{Influence of the chosen $L^p$ loss on the instance segmentation mask AP COCO metrics on the MS COCO dataset after $160$k steps. Higher is better.}{
        $L^1$ & $19.5 \pm 0.28$ & $36.6 \pm 0.41$ & $18.9 \pm 0.41$ & $8.2 \pm 0.30$ & $21.0 \pm 0.32$ & $27.7 \pm 0.5$ & $11\%$\\
        $L^2$ & $\mathbf{20.2 \pm 0.17}$ & $\mathbf{37.5 \pm 0.37}$ & $20.0 \pm 0.07$ & $\mathbf{8.8 \pm 0.27}$ & $\mathbf{21.9 \pm 0.18}$ & $\mathbf{28.9 \pm 0.30}$ & $\mathbf{29\%}$\\
        $L^3$ & $\mathbf{20.2 \pm 0.20}$ & $37.0 \pm 0.41$ & $\mathbf{20.1 \pm 0.22}$ & $8.6 \pm 0.14$ & $21.8 \pm 0.24$ & $28.5 \pm 0.53$ & $22\%$\\
        $L^4$ & $17.8 \pm 0.13$ & $33.5 \pm 0.12$ & $17.4 \pm 0.13$ & $7.6 \pm 0.12$ & $19.3 \pm 0.23$ & $24.7 \pm 0.24$ & $0\%$\\
        \hline
        Baseline  & $18.8 \pm 0.14$ & $36.5 \pm 0.24$ & $17.8 \pm 0.13$ & $8.0 \pm 0.21$ & $20.4 \pm 0.29$ & $26.6 \pm 0.24$ & $-$\\
    }
    
    The two losses $L_{MRCNN}$ and $L_{Mask}$ are displayed in Figure~\ref{fig:influence_lp}. While a higher value for $p$ causes the mask loss to decrease, it also increases the overall loss. The metrics obtained in these experiments are listed in Table~\ref{tab:experiment_lp_norm}. Overall, choosing the $L^2$ loss appears to be the best choice, as it yields the best results on the COCO metrics.
    
\subsection{Influence of Weighting Factor on Edge Agreement Loss}\label{sec:lp_loss_wf_factor}
    By choosing a higher value of $p$ for the $L^p$ Edge Agreement Loss, the loss becomes steeper and yields higher values for wrongly predicted masks. This increment also implies a higher relative importance of the Edge Agreement Loss compared to the other loss functions in the sum of the total loss which usually stay in the range $[0, 1]$. 
    
    In these trainings we used the Sobel edge detection filter.
    
    To examine the influence of the relative importance of the new Edge Agreement Loss to the other losses, we include a factor $\alpha$ which scales the Edge Agreement Loss. We test its influence on the usage of the $L^2$ and $L^4$ losses to investigate the impact of the Mask Edge Loss on the total loss. For this comparison all trainings are performed only once and up to $120$k steps instead of $160$k steps, as already after $120$k steps a clear trend has been recognizable.
    
    \begin{figure*}[t!]
        \centering 
        \hfill
        \begin{minipage}{0.66\textwidth}
            \begin{subfigure}{.5\linewidth}
                \centering
                \includegraphics{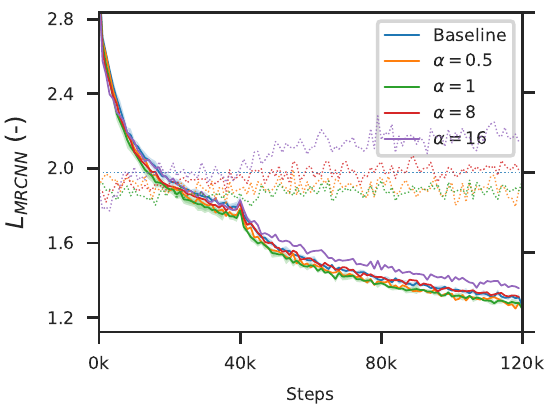}
                \caption{Different values for $\alpha$ for $L^2$ loss}\label{fig:wf_comparison_l2}
            \end{subfigure}
            \begin{subfigure}{.5\linewidth}
                \centering
                \includegraphics{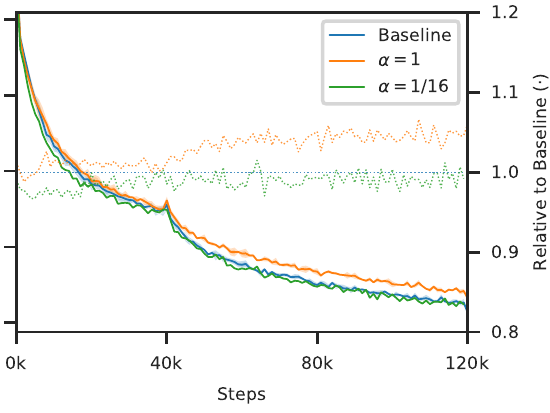}
                \caption{Different values for $\alpha$ for $L^4$ loss}\label{fig:wf_comparison_l4}
            \end{subfigure}
            
            \bigskip
            \vspace*{-4mm}
            \begin{subfigure}{.5\linewidth}
                \centering
                \includegraphics{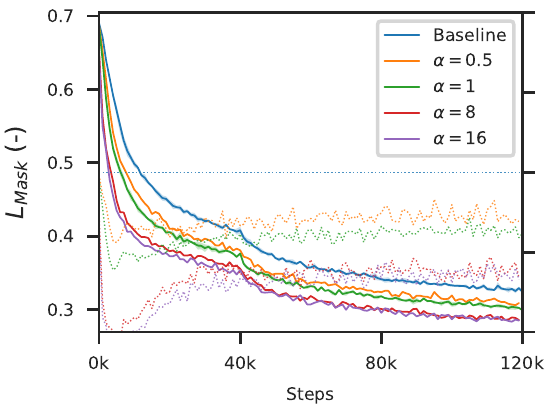}
            \end{subfigure}
            \begin{subfigure}{.5\linewidth}
                \centering
                \includegraphics{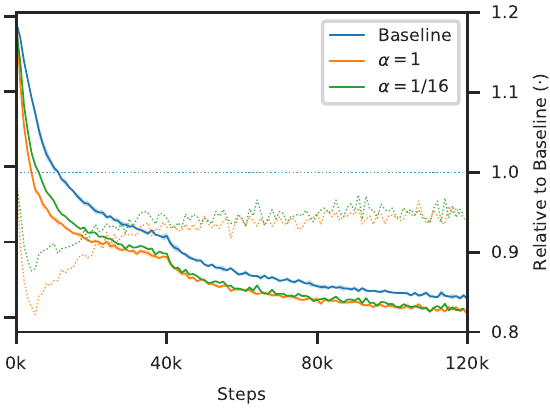}
            \end{subfigure}
        \end{minipage}%
        \quad
        \begin{minipage}{.30\textwidth}
            \centering
            \label{fig:wf_comparison}
             \caption{Influence of the weighting factor $\alpha$ on the behavior of the Edge Agreement Loss on the MS COCO dataset. The first row displays again the original total loss of Mask R-CNN $L_{MRCNN}$ while the second row displays only the mask loss $L_ {Mask}$. The first column  shows the loss trajectory for different alpha values using the $L^2$ loss whereas the second column shows the influence on the $L^4$ loss. Best viewed in color}
         \end{minipage}
    \end{figure*}
    
    Figure~\ref{fig:wf_comparison_l2} shows the Mask R-CNN loss, while the Edge Agreement Loss is scaled by $\alpha \in \{0.5, 1, 8, 16\}$. The Mask R-CNN loss increases with higher weight factor, despite faster decreasing Mask Loss, the other loss terms remain higher. In fact, the $L^2$ loss with weight factor 1.0 already appears to be a good trade-off between enforcing better predicted masks and optimizing the other objectives of the network.
    
    The $L^4$ loss yields high values compared to the $L^2$ loss. Therefore, we scale it by $\alpha = 1/16$, which is approximately the ratio of Edge Agreement Loss between using $L^2$ and $L^4$ in the first few steps. The training progression for the Mask R-CNN loss is shown in Figure~\ref{fig:wf_comparison_l4}. Reducing the Edge Agreement Loss improves training significantly, making the loss stay below the Baseline for most of the steps.  

\subsection{Influence of Mask Size}\label{sec:influence_of_mask_size}
    We investigate the influence of the size of the predicted mask on the performance of the Edge Agreement Head (see Table~\ref{tab:mask_size}). For this we compare the performance of models trained for the original mask size of $28 \times 28$ and for an increased size of $56 \times 56$. The models were trained following the same training schedule as outlined in section \ref{sec:implementation}. For both mask sizes we observe a clear increase in performance when using the Edge Agreement Head as an auxiliary loss. However, we find that the overall performance of the model trained at mask size $56 \times 56$ is worse than the model trained with mask size $28 \times 28$. We hypothesize that this is the case, because predicting twice the resolution would require longer training.
    
    \cocometricstableshape[AP]{tab:mask_size}{Influence of the size of the predicted and ground-truth masks on the Edge Agreement Head. Shown are the instance segmentation mask AP COCO metrics on the MS COCO dataset.}{
        \multirow{2}{*}{$28 \times 28$}
        & Baseline  & $18.8 \pm 0.14$ & $36.5 \pm 0.24$ & $17.8 \pm 0.13$ & $8.0 \pm 0.21$ & $20.4 \pm 0.29$ & $26.6 \pm 0.24$  & $-$\\  
        & Ours & $\mathbf{20.2 \pm 0.17}$ & $\mathbf{37.5 \pm 0.37}$ & $\mathbf{20.0 \pm 0.07}$ & $\mathbf{8.8 \pm 0.27}$ & $\mathbf{21.9 \pm 0.18}$ & $\mathbf{28.9 \pm 0.30}$ & $\mathbf{29}\%$\\
        \hline
        \multirow{2}{*}{$56 \times 56$}
        & Baseline & $18.0 \pm 0.23$ & $35.0 \pm 0.30$ & $17.0 \pm 0.37$ & $7.6 \pm 0.24$ & $19.3 \pm 0.28$ & $25.5 \pm 0.31$  & $-$\\  
        & Ours & $19.3 \pm 0.03$ & $36.0 \pm 0.07$ & $19.0 \pm 0.07$ & $8.1 \pm 0.13$ & $21.0 \pm 0.07$ & $27.6 \pm 0.19$ & $20\%$
    }

\subsection{Longer training} \label{sec:longer_training}
    \cocometricstable[AP]{tab:longer_training}{Comparison of the instance segmentation mask AP COCO metrics on the MS COCO dataset of our best performing model with the baseline after an extended training duration. The best performing model uses the Edge Agreement Head with Sobel edge detection filter and $L^2$ Edge Agreement Loss.}{
        Ours $160$k steps & $20.2 \pm 0.17$ & $37.5 \pm 0.37$ & $20.0 \pm 0.07$ & $8.8 \pm 0.27$ & $21.9 \pm 0.18$ & $28.9 \pm 0.30$\\
        Ours $320$k steps & $21.3$ & $38.7$ & $21.1$ & $8.9$ & $23.2$ & $30.0$ \\
        Ours $640$k steps & $\mathbf{22.7}$ & $\mathbf{41.0}$ & $\mathbf{23.1}$ & $\mathbf{10.2}$ & $\mathbf{24.6}$ & $\mathbf{32.0}$ \\
        \hline
        Baseline $160$k steps & $18.8 \pm 0.14$ & $36.5 \pm 0.24$ & $17.8 \pm 0.13$ & $8.0 \pm 0.21$ & $20.4 \pm 0.29$ & $26.6 \pm 0.24$ \\  
        Baseline $320$k steps & $20.0$ & $38.5$ & $19.1$ & $8.6$ & $21.6$ & $28.2$\\
        Baseline $640$k steps & $21.5$ & $40.5$ & $20.8$ & $9.0$ & $22.9$ & $30.7$\\
    }
    To measure the effect of the Edge Agreement Head after longer training, we increased the number of steps previously used ($320$k and $640$k steps rather than $160$k). In this case the last step of the training schedule, in which all layers are trained, was extended from $40$k steps to $200$k and $520$k steps (a total of $320$k and $640$k training steps respectively). The results are shown in Table~\ref{tab:longer_training}.
    
    All metrics improved as expected when trained for additional steps. Interestingly, in most metrics the Mask R-CNN model trained with Edge Agreement Head trained for $160$k steps was not only superior to the baseline trained for $160$k steps but also to the one trained $320$k steps. No significant influence of the Edge Agreement Loss on losses other than the Mask Loss is observed. We notice that the difference in the Mask Loss between a baseline Mask R-CNN and one trained with Edge Agreement Loss remains constant with later training steps. This was contrary to our own intuition that the Edge Agreement Loss would primarily be helpful early in training. It was expected that the Mask Loss of the two models would approach each other, but this was not found to be the case. We conclude that the Edge Agreement Head is not only useful early on in training, but can guide the training even in later training steps and change the point of convergence.
    
    It should be noted that our results on the MS COCO dataset \citep{lin2014microsoft} are significantly lower than the results reported by \cite{he2017mask}. Firstly, we are not using the official implementation of Mask R-CNN made available in the Detectron \citep{Detectron2018}, but an independent implementation which reported lower results of their pretrained models \citep{matterport_maskrcnn_2017}. Secondly, we use a batch size of 2, while Mask R-CNN used an effective batch size of 16. We argue that this does not hurt the generality of our method. 

\subsection{Other experiments}\label{sec:other_experiments}
    The configuration of the Edge Agreement Loss we describe above was found to have the optimal impact on the training. We tried a variety of modifications which showed either no effect or had a negative impact on the training. 
    
    \subsubsection{Smoothing of ground-truth or predicted Masks}
    Figure~\ref{fig:mask_gt_prediction_blurred_gt_loss} illustrates how the $L^2$ loss between the edge maps, calculated as mentioned above, does not only contain important information but even possibly distracting information: the images in the rows for Sobel and Laplace depict a high error rate which is not limited to areas in which the person has not been segmented but also occurs around the entire boundary. This is mainly caused by the fact, that the ground-truth mask is binary and the mask branch's output is continuous and shows often smooth transitions.
    
    \begin{figure}[h]
        \centering
        \vspace*{-2mm}
        \includegraphics[width=3.4in]{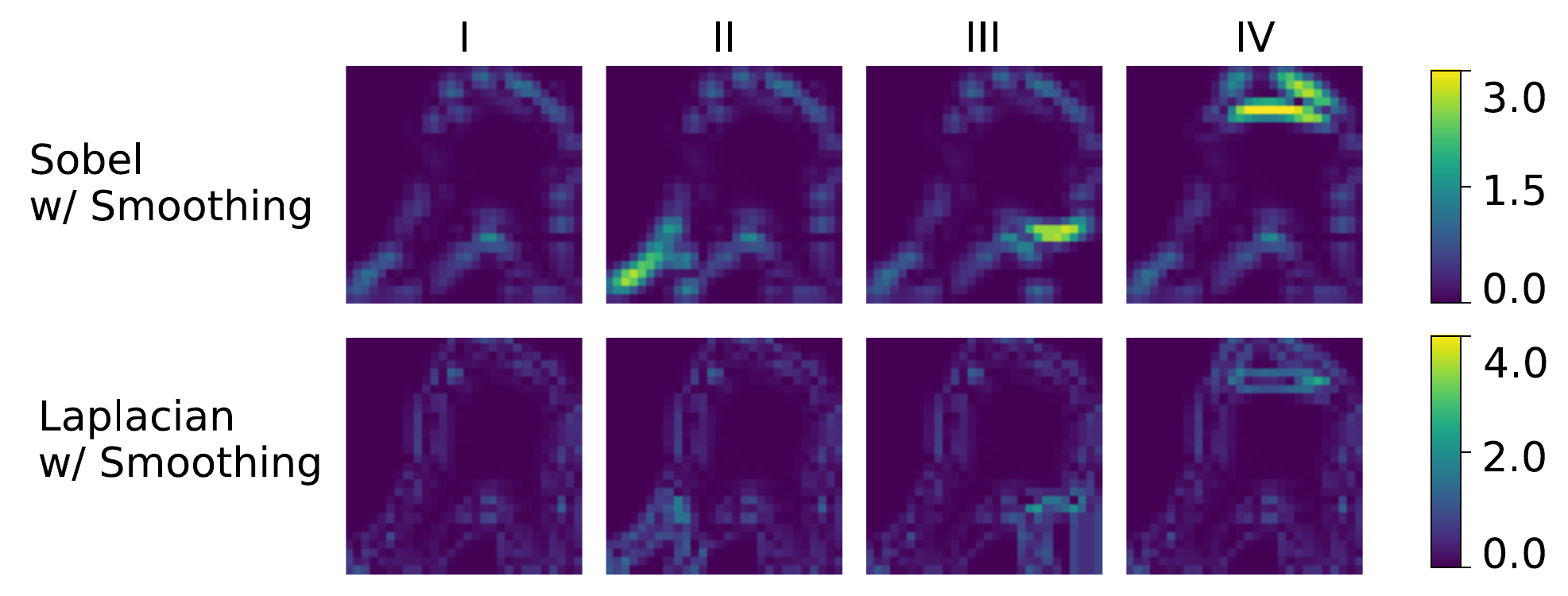}
        \vspace*{-3mm}
        \caption{$L^2$ errors for the four example predictions and the different filters calculated on the Gaussian smoothed ground truth. Each column $I$ to $IV$ corresponds to one of the examples in Figure~\ref{fig:mask_gt_prediction_overview}. The first row shows the $L^2$ error based on the Sobel filter magnitude, while for the second row the Laplace filter is used. While the losses calculated on the default ground truth (see Figure~\ref{fig:mask_gt_prediction_blurred_gt_loss}) do not respond strongly to missing areas, the losses on the smoothed ground truth are particularly pronounced in these areas.}
        \label{fig:mask_gt_prediction_blurred_gt_loss_smoothed}
    \end{figure}

    To overcome this problem, we add an additional step in our branch which performs Gaussian smoothing on the ground-truth mask, yielding a smooth version of the binary ground truth. For this we use an approximate $3 \times 3$ Gaussian kernel. The calculated $L^p$ distance using this head proposal is shown in Figure~\ref{fig:mask_gt_prediction_blurred_gt_loss_smoothed}. Notably, the loss calculated on the smoothed ground truth focuses particularly on areas with missing parts, while the Edge Agreement Loss on the default ground truth has a higher value on the overall mask boundary.
    
    Contrary to expectation, the Sobel filter was more effective when used without smoothing the ground truth, as the Mask R-CNN training loss fell faster and was lower in this case, as shown in Figure~\ref{fig:influence_smoothing}. Particularly the Mask Loss  during training is lower.
    
    The results obtained contradict the theoretical considerations of the possible benefit to smoothing the ground truth. The smoothing of the ground truth was designed to ignore minor mistakes at the boundary of an almost perfectly predicted mask, but only focusing on major mistakes. Apparently, the network does not only profit from highlighting the most crucial mistakes in the predicted masks, but rather from all mistakes done.
    
    We investigated whether smoothing both the ground truth and the predicted mask or only the predicted mask would help the network during training. The reasoning for this was that it could be beneficial to penalize the network less for pixel accurate mask and more for the general shape. Since the instance boundaries become much wider due to the smoothing, the Edge Agreement Loss becomes less sensitive to small spatial displacements. However, we found this modification to have a negative impact on the training.
    
    \subsubsection{Balancing Losses}
    
    As discussed in section~\ref{sec:lp_loss_wf_factor}, the magnitude of the Edge Agreement Loss appears to have a high influence on the $L_{MRCNN}$ loss. In an attempt to balance the loss terms we tried homoscedastic task uncertainty as proposed by \cite{kendall2017multi}. Our approach was to weigh all the loss terms including the Edge Agreement Loss. However, the results were consistently worse than the baseline and therefore not included in this paper.
    
    \subsubsection{Alternative Edge Loss Definitions}
    Furthermore, we tried to weigh the cross entropy mask loss $L_{Mask}$ with the Edge Agreement Loss. Two different formulations for this weighted cross entropy loss were tried out, which can be expressed as
    \begin{align*}
        L_{Edge} &= L_{Mask-PW}\cdot L_{Edge-PW} \\[-1em]
        \intertext{\centering or} \\[-3em]
        L_{Edge} &= L_{Mask-PW} \cdot \exp \left(L_{Edge-PW} / 4\right),
    \end{align*}
    using $L_{Mask-pw}$ and $L_{Edge-pw}$ to denote the pixel-wise Mask Loss and pixel-wise Edge Agreement Loss respectively. 
    For both formulations the results were identical with the more concise formulation of the Edge Agreement Loss that we used in the rest of the paper.

    In addition, when using the Sobel filter, we did not solely consider the horizontal and vertical image gradient but also the gradient's magnitude for calculating the $L^p$ Edge Agreement Loss. No improvement compared to not including the magnitude was found. Therefore, it was not used in the rest of the paper.

\subsection{Cityscapes}
    To verify that our findings could be reproduced on a different dataset, we trained our models on the Cityscapes dataset \citep{cordts2016cityscapes} with the Edge Agreement Head with Mask shape $28 \times 28$ and $56 \times 56$ pixels (see Table~\ref{tab:mask_size_cityscapes}). In contrast to MS COCO, the annotations in Cityscapes have much finer details. We followed the training schedule of the authors of Mask R-CNN \citep{he2017mask}, but instead of using an effective batch size of $8$ we used a reduced size of $4$. The findings are easily compared to Table \ref{tab:mask_size}, since we see very similar trends. Training with the Edge Agreement Head is demonstrated to be beneficial since it consistently outperforms the respective baseline. Increasing the predicted mask size leads to an overall reduction in the accuracy of the predicted mask for the baseline experiments, but this could potentially be remedied by longer training, to account for the higher number of parameters in the additional layer of the mask branch. 
    \cocometricstableshape[AP]{tab:mask_size_cityscapes}{Influence of the size of the predicted and ground-truth masks on the Edge Agreement Head with Sobel and an $L^2$ loss. Shown are the instance segmentation mask AP COCO metrics on the Cityscapes dataset.}{
        \multirow{2}{*}{$28 \times 28$}
        & Baseline  & $15.6 \pm 0.84$ & $33.5 \pm 1.8$ & $12.43 \pm 1.03$ & $2.2 \pm 0.25$ & $11.2 \pm 0.43$ & $25.6 \pm 1.21$  & $-$\\  
        & Ours & $17.7 \pm 0.49$ & $\mathbf{36.1 \pm 0.78}$ & $14.5 \pm 0.74$ & $\mathbf{3.0 \pm 0.10}$ & $12.5 \pm 0.81$ & $\mathbf{33.3 \pm 4.54}$  & $\mathbf{20}\%$\\
        \hline
        \multirow{2}{*}{$56 \times 56$}
        & Baseline & $14.9 \pm 0.40$ & $31.5 \pm 0.22$ & $11.9 \pm 0.63$ & $2.1 \pm 0.11$ & $10.2 \pm 0.97$ & $24.4 \pm 0.82$  & $-$\\  
        & Ours & $\mathbf{18.0 \pm 1.09}$ & $35.0 \pm 2.65$ & $\mathbf{15.05 \pm 0.99}$ & $1.8 \pm 0.04$ & $\mathbf{12.7 \pm 0.64}$ & $31.5 \pm 0.297$  & $\mathbf{20}\%$\\
    }
    
\section{Conclusion}
    In this paper we have analyzed the behavior of Mask R-CNN networks during early training steps. By inspecting the predicted masks of the mask branch, we recognized that these often have blurry boundaries which do not follow sharp and fine contours of the original masks. To reduce this symptom, we successfully introduced a parameter free network head, the Edge Agreement Head. This head uses classical edge detection filters applied on the instance masks to calculate a $L^p$ loss between the predicted and ground-truth mask contours.
    
    By including the new Edge Agreement Loss in the training, we achieved a relative performance increment of 8.1\% averaged over all the MS COCO metrics after a fixed number of $160$k training steps.
    
    The ablation studies performed showed that the Sobel filter yields a better performance than the Laplace filter. Beyond expectations, the proposed smoothing of the ground-truth mask did not improve but hinder the performance. Out of all losses examined the often-used $L^2$ loss performs the best. 
    
    When trained longer, the difference in Mask Loss between a baseline Mask R-CNN and one with Edge Agreement Head persists, demonstrating the effectiveness of the additional loss not only early during training but also during later steps.

    Finally, we demonstrated that the Edge Agreement Head is beneficial on Cityscapes, a dataset with much finer ground-truth masks.
    
\section{Future work}
    The idea to enforce edge agreement in predicted semantic segmentation could be applied to scene segmentation for example on the DeepLab architecture \citep{Chen2018DeepLabSI} or the U-Net architecture \citep{ronneberger2015u, kohl2018probabilistic}. Monocular depth estimation could also potentially be enhanced by encouraging the predicted depth map to have comparable gradients to the ground-truth depth map image gradients.
    
    Furthermore, balancing the different individual losses contained in the total loss by introducing new scaling variables might be a necessary step to further increase the training speed. Instead of introducing new static hyperparameters for the multi-task loss one could modify the gradients like \cite{Chen2018GradNormGN}.
    
    As the Edge Agreement Loss accelerates the training of the Mask Head, it enables Mask R-CNN to be used more easily with sparse labels for object instance masks. This allows new training strategies for new datasets, e.g. one could mix a few hand segmented frames with datasets containing only object bounding boxes, such as PASCAL VOC \citep{Everingham2010} or the more recent Open Images dataset \citep{OpenImages2}.

\section*{Acknowledgments}
    This research has been started during an internship at BMW Car IT GmbH in Ulm. The researchers would like to thank the company for the financial support and for the resources offered. Furthermore, the authors sincerely thank Krassimir Valev and Stefan M\"uller from BMW Car IT GmbH for their help in organizing this publication.

\appendix
\section*{References}
\bibliographystyle{apalike}
\bibliography{library}

\section*{Authors}

    \authorbio{Roland S. Zimmermann}{received his B.Sc. degree with distinction in Physics from Georg-August University of G\"ottingen, G\"ottingen, Germany in 2017 where he is currently pursuing his M.Sc. in collaboration with the University of T\"ubingen, T\"ubingen, Germany. His research interests lie in the areas of computer vision and (adversarial) robustness of neural networks.}{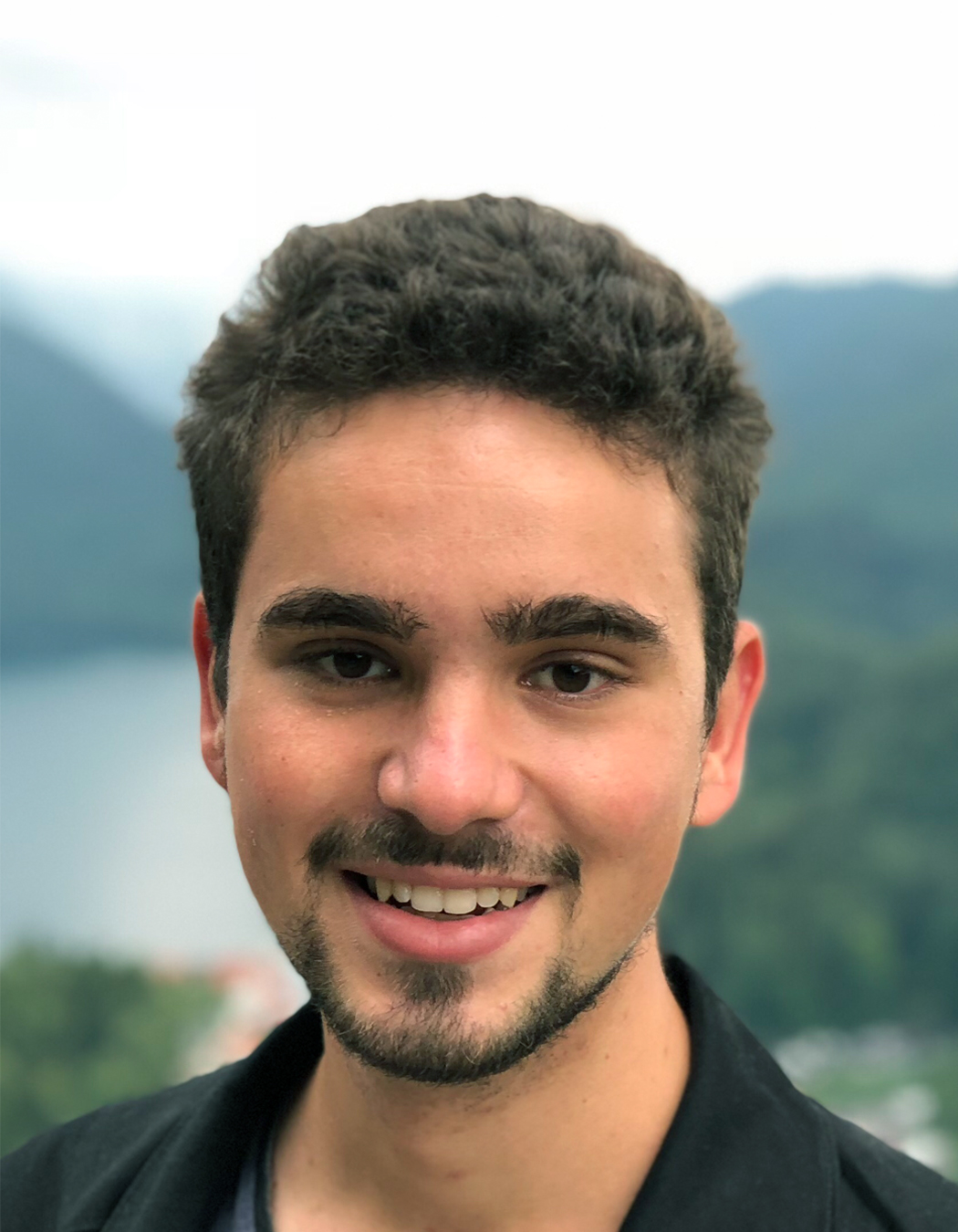}
    
    \vspace{1cm}
    
    \authorbio{Julien N. Siems}{received his B.Sc. degree in Computer Science from TU Dresden, Dresden, Germany in 2017. He is currently pursuing his M.Sc. at the Albert Ludwig University of Freiburg, Freiburg, Germany. His research interests lie in the areas of computer vision and neural architecture search.}{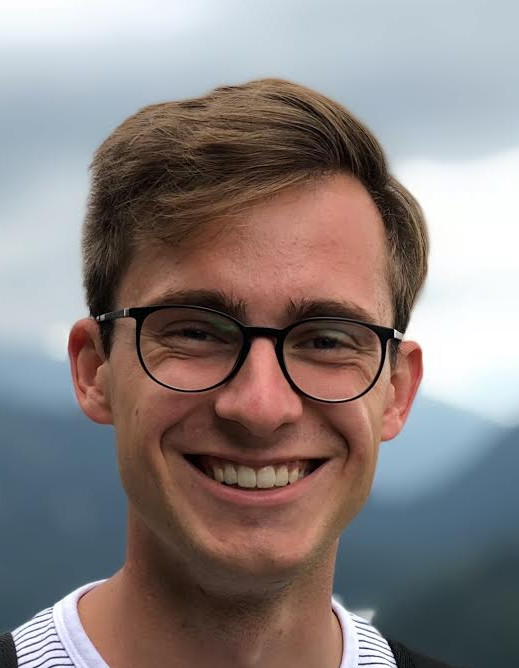}

\end{document}